\documentclass{ifacconf}

\usepackage{graphicx}      
\usepackage{natbib}        
\usepackage{amsmath}
\usepackage{amssymb}
\usepackage{array}
\usepackage{subfig}
\usepackage[width=0.9\textwidth]{caption}
\usepackage{url}

\DeclareMathOperator*{\argmax}{arg\,max}

\newcommand*{\tran}{^{\mkern-1.5mu\mathsf{T}}}

\begin{document}
\begin{frontmatter}

\title{Deep Reinforcement Learning with Embedded LQR Controllers\thanksref{footnoteinfo}} 

\thanks[footnoteinfo]{\copyright 2020 the authors. This work has been accepted to IFAC for publication under a Creative Commons Licence CC-BY-NC-ND. This research was partially supported by the Brazilian National Council for Scientific and Technological Development (CNPq), grant number 304980/2018-8.}

\author[First]{Wouter Caarls} 

\address[First]{Pontifical Catholic University, Rio de Janeiro, RJ 22451-900 Brazil (e-mail: wouter@ele.puc-rio.br).}

\begin{abstract} 
Reinforcement learning is a model-free optimal control method that optimizes a control policy through direct interaction with the environment. For reaching tasks that end in regulation, popular discrete-action methods are not well suited due to chattering in the goal state. We compare three different ways to solve this problem through combining reinforcement learning with classical LQR control. In particular, we introduce a method that integrates LQR control into the action set, allowing generalization and avoiding fixing the computed control in the replay memory if it is based on learned dynamics. We also embed LQR control into a continuous-action method. In all cases, we show that adding LQR control can improve performance, although the effect is more profound if it can be used to augment a discrete action set.
\end{abstract}


\end{frontmatter}

\section{Introduction}
Reinforcement Learning (RL,~\cite{sutton2018}) is an optimal control method that estimates control policies from direct interaction with a (real or simulated) environment. It has been sucessfully applied in such diverse areas as robotics~(\cite{akkaya2019solving}), game playing~(\cite{vinyals2019grandmaster}), and medication dosing~(\cite{nemati2016optimal}). RL methods can be divided in two broad categories: those that directly optimize the parameters of an explicit control policy (called policy search methods), and those that estimate the reward-to-go (value) function and implicitly derive a control policy from that (value-based methods).

Value-based methods are better for tasks that inherently contain many local optima (``maze-like tasks''). However, they struggle in problems with continuous action spaces such as robotics, because the policy is defined as taking the action with the highest reward-to-go, which is usually implemented by iteration over a discrete set of possible actions. This is especially relevant for regulation or reaching tasks, because a controller using discrete actions can never maintain a stable dynamic position.The resulting chattering is undesirable and can lead to system damage (\cite{mpc:pada}).

One solution is to use a lower-level (continuous-action) stabilizing controller at the goal state, while reinforcement learning optimizes the trajectory towards that goal. This can for example be implemented by defining a ``capture region" in which the stabilizing controller is active~(\cite{Randlov:2000:CRL:645529.657804}), or by making it one of the possible discrete actions~(\cite{abramova2019rloc}). The first case requires manual fine-tuning of this region, while the second requires learning the value of a new action. In this paper, we propose using the stabilizing controller's action one of the sampling points of the discrete action set, thus taking advantage of the action-space generalization provided by modern deep reinforcement learning methods.

In addition, we investigate the use of the same approach in an actor-critic setting. Actor-critic methods are a hybrid of policy-based and value-based RL methods. They use a value function to inform the adaptation of an explicitly represented policy. Although they do not suffer from a discrete action set, the intuition is that the inclusion of the action suggested by a (model-based) stabilizing controller could increase the learning speed, as it provides near-optimal actions in at least part of the state space.

The paper is organized as follows. Sections~\ref{sec:theory} and~\ref{sec:methods} introduce the basic theory and our specific methods of combining RL with LQR control, respectively. Sections~\ref{sec:experiments} and~\ref{sec:results} present the experiments, results and their analysis. Finally, section~\ref{sec:conclusion} concludes the paper and discusses future work.

\section{Theory}
\label{sec:theory}

\subsection{Reinforcement Learning}

A Markov Decision Process (MDP) is a 4-tuple $<\mathcal{S}, \mathcal{A}, T, R>$ where $\mathcal{S}$ is the (possibly continuous) state space, $\mathcal{A}$ the (possibly continuous) action space, $T: \mathcal{S}, \mathcal{A}, \mathcal{S} \rightarrow \mathbb{R}$ a function specifying the state transition probabilities and $R: \mathcal{A}, \mathcal{S} \rightarrow \mathbb{R}$ a reward function. The goal of reinforcement learning is to find a control policy $\pi: \mathcal{S} \rightarrow\mathcal{A}$ that maximizes, for every state $s\in\mathcal{S}$ the expected \emph{discounted return} $\mathfrak{R_t}$ of executing that policy starting from $s$ at the current timestep $t$:

\begin{equation}
\mathfrak{R}_t = \sum_{k=0}^\infty \gamma^k r_{t+k}
\end{equation}

where $r_k$ is the reward received at timestep $k$ and $\gamma \in [0, 1]$ is a discount factor that determines the optimization horizon.

In value-based reinforcement learning, the expected returns are stored in a state-value function $V^\pi: \mathcal{S} \rightarrow \mathbb{R}$, or an action-value function $Q^\pi: \mathcal{S}, \mathcal{A} \rightarrow \mathbb{R}$. In the latter case, the stored values $Q(s, a)$ indicate the return of taking action $a$ in state $s$ and following $\pi$ afterwards. The optimal policy $\pi^*$ always takes the action with the highest expected return, and its value functions are therefore related:

\begin{align}
\label{eq:pi}
\pi^*(s) &= \argmax_{a\in\mathcal{A}} Q^*(s, a)\\
\label{eq:v}
V^*(s) &= \max_{a\in\mathcal{A}} Q^*(s, a).
\end{align}

$Q^*$ is the unique solution to the Bellman optimality equation

\begin{equation}
\label{eq:bellman}
Q^*(s, a) = \sum_{s'}T(s, a, s')\left(R(a, s') + \gamma V^*(s')\right).
\end{equation}

Note that it is not necessary to define $V^*$ in terms of $Q^*$, as Eqs.~\ref{eq:v}  and~\ref{eq:bellman} may be combined into a single equation. However, in that case the transition probabilities $T$ must be known in order to derive $\pi^*$ in Eq.~\ref{eq:pi}.

We consider model-free reinforcement learning, in which the transition probabilities $T$ are unknown. In that case, the value functions are usually estimated by sampling experienced \emph{transitions} $(s_t, a_t, r_t, s_{t+1})$. Each transition gives a sample of the Bellman optimality equation (\ref{eq:bellman}):

\begin{align}
Q(s_t, a_t) &= r_t + \gamma V(s_{t+1})\\
\label{eq:q}
&=r_t + \gamma \argmax_{a' \in\mathcal{A}} Q(s_{t+1}, a')
\end{align}

\subsubsection{Deep Q learning}

In deep Q learning (DQN,~\cite{deeprl}), the sampled transitions, gathered under an exploratory policy based on $\pi$, are stored in a \emph{replay memory} which is used to approximate the action-value function with a deep neural network $\hat{Q}(s, a|\theta^Q)$ with parameters $\theta^Q$. Iteratively (generally after every new transition), a \emph{mini-batch} of $N$ transitions $(s_i, a_i, r_i, s'_i)$ is sampled randomly from this replay memory and used to minimize the $L_2$ loss between the left-hand and right-hand sides of Eq~\ref{eq:q}:

\begin{equation}
\label{eq:loss}
\mathcal{L} = \sum_i\left\|\hat{Q}(s_i, a_i|\theta^{Q}) - \left(r_i + \gamma \argmax_{a' \in\mathcal{A}} \hat{Q}(s'_i, a'|\theta^{Q'})\right)\right\|_2.
\end{equation}

Note the use of different weights $\theta^{Q'}$ in the right-hand side, used to stabilize the learning. These \emph{target weights} are updated periodically from the learned weights $\theta^Q$. The exploratory policy is usually $\epsilon$-greedy with respect to Eq.~\ref{eq:pi}:

\begin{align}
\label{eq:egreedy}
\pi^\epsilon(s) = \left\{ \begin{array}{ll}
\argmax_{a\in\mathcal{A}}Q(s, a)&\text{with probability $\epsilon$}\\
\mathrm{random}(\mathcal{A})&\text{otherwise}
\end{array}\right. .
\end{align}

\subsubsection{Deep Deterministic Policy Gradient}

Actor-critic algorithms represent $\pi$ explicitly instead of deriving it from $Q$. In the deep deterministic policy gradient algorithm (DDPG, ~\cite{lillicrap:ddpg}), the $Q$ function is approximated as in deep Q learning, but the $\argmax$ in Eq.~\ref{eq:loss} is replaced by the action given by a separate policy network $\mu(s|\theta^\mu)$ trained to be the mean of a Gaussian policy

\begin{equation}
\label{eq:gauss}
\pi(s|\theta^\mu) = \mathcal{N}(\mu(s|\theta^\mu), \sigma)
\end{equation}

where $\sigma$ is the standard deviation of the exploration distribution. After every $Q$ update, the weights $\theta^\mu$ of the policy network are updated in the direction of

\begin{multline}
\mathbb{E}\left[\nabla_{\theta^\mu}Q(s, a|\theta^Q)|_{s=s_t,a=\mu(s_t|\theta^\mu)}\right] \approx\\
\frac{1}{N}\sum\nabla_a Q(s, a|\theta^Q)|_{s=s_i,a=\mu(s_i|\theta^\mu)} \nabla_{\theta^\mu}\mu(s|\theta^\mu)|_{s=s_i} .
\end{multline}

As in deep Q learning, separate target networks with weights $\theta^{Q'}$ and $\theta^{\mu'}$ are used to calculate the right hand side of Eq.~\ref{eq:loss}. However, instead of periodic updates, they are updated continuously from $\theta^{Q}$ and $\theta^{\mu}$ using a moving average filter. In addition, the Gaussian policy is usually replaced by one that has time-correlated noise to improve exploration~(\cite{lillicrap:ddpg}).

\subsection{Locally Linear Regression}

We use locally linear regression (LLR,~\cite{atkeson1997a}) to approximate the dynamics around the goal state $s_\mathrm{d}$. LLR is well-suited to model approximation for LQR because the linear model it estimates can be directly split into the $A$ and $B$ matrices required by LQR control. In LLR, all transitions $(s_i, a_i, s'_i)$ are stored and a linear model $X$ is fit around the $K$ nearest neighbors of the query point $s_\mathrm{d}$

\begin{equation}
N_\mathrm{I} = \left[\begin{matrix}
\overline{s_1}, a_1, 1\\
\overline{s_2}, a_2, 1\\
\cdots\\
\overline{s_K}, a_K, 1\\
\end{matrix}\right],
N_\mathrm{O} = \left[\begin{matrix}
\Delta s_1\\
\Delta s_2\\
\cdots\\
\Delta s_K\\
\end{matrix}\right],
\end{equation}

where $\overline{s_i}=s_i-s_\mathrm{d}$ and $\Delta s_i = s'_i - s_i$, by solving

\begin{equation}
(N_\mathrm{I}\tran N_\mathrm{I})X = N_\mathrm{I}\tran N_\mathrm{O}
\label{eq:chol}
\end{equation}

using the Cholesky decomposition with Tikhonov regularization. The $A$ and $B$ matrices of the discrete-time system 

\begin{equation}
\label{eq:system}
s_{t+1} = As_t + Ba_t + E
\end{equation}

are then defined as follows:

\begin{align}
A &= \overline{X}_{1:|\mathcal{S}|, 1:|\mathcal{S}|}\\
B &= \overline{X}_{1:|\mathcal{S}|, |\mathcal{S}|+1:|\mathcal{S}|+|\mathcal{A}|},
\end{align}

where $\overline{X} = X + I$, and $|\mathcal{S}|$ and $|\mathcal{A}|$ are the dimensionalities of $\mathcal{S}$ and $\mathcal{A}$, respectively.

\subsection{Linear Quadratic Regulator}

The Linear Quadratic Regulator is an optimal solution to the Bellman equation (\ref{eq:bellman}) for linear systems with quadratic rewards~(\cite{mehrmann1991autonomous}), in our case

\begin{equation}
R(s, a) = -\left(\overline{s}\tran C \overline{s}+a\tran D a \right),
\end{equation}

where $C$ and $D$ are diagonal matrices specifying the costs for deviating from the goal state $s_\mathrm{d}$ and zero action, respectively. The solution is computed by solving the discrete time algebraic Riccati equation~(\cite{benner2003solving})

\begin{equation}
P = A\tran PA  - \left(A\tran PB\right)\left(D + B\tran PB\right)^{-1}\left(B\tran PA\right)+C
\end{equation}

after which the control is given by 

\begin{align}
a_t &= -F \overline{s_t}\\
F &= \left(D + B\tran PB\right)^{-1}\left(B\tran PA\right).
\end{align}

To find the steady-state feedforward action $a_\mathrm{ff}$ to cancel the disregarded $E$ term in Eq.~\ref{eq:system}, we solve

\begin{equation}
Ba_\mathrm{ff} = As_{t+1}|_{s_t=s_\mathrm{d},a_t=0} - s_\mathrm{d}
\end{equation}

using the singular value decomposition and add it to the regulator output.

\section{Methods}
\label{sec:methods}

We compare three different ways of integrating LQR control with reinforcement learning: LQR capture, LQR action and integrated LQR action.

\subsection{LQR Capture}

In LQR capture, the system is controlled by the reinforcement learning agent except in a region around the goal state, in which case control is taken over by the LQR controller~(\cite{Randlov:2000:CRL:645529.657804}). From the point of view of the RL agent, this turns the system into a semi-MDP (sMDP,~\cite{sutton1999between}), where the action that led to the capture region is temporally extended until the system leaves the capture region, or until the end of the episode. In sMDPs, the target in Eq.~\ref{eq:q} is replaced by 

\begin{equation}
\label{eq:smdp}
Q(s_t, a_t) = \sum_{k=0}^{\Delta t-1}\gamma^k r_{t+k} + \gamma^{\Delta t} \argmax_{a'\in\mathcal{A}} Q(s_{t+1}, a'),
\end{equation}

where $\Delta t$ is the number of time steps taken by action $a_t$. If the episode terminates with the system still within the capture region, we treat the state as \emph{terminal absorbing}~(\cite{sutton2018}), with the reward given by Eq.~\ref{eq:smdp} and $\Delta t$ equal to the number of steps the system remained in the capture region before the episode terminated. As such, the agent learns to enter the capture region in such a way as to maximize the performance of the LQR controller, instead of just entering the capture region itself such as in~(\cite{Randlov:2000:CRL:645529.657804}).

\subsection{LQR Action}

LQR capture requires the system architect to specify a capture region. If this region is not chosen optimally, the resulting controller will not be optimal. To avoid manual specification of the capture region, we may instead allow the reinforcement learning controller to select a special action which applies the action suggested by the LQR controller. Although this requires learning the return of this new action, it allows the controller to stabilize a continuous system.

For the LQR action algorithm, the action set is thus expanded to

\begin{equation}
\mathcal{A}^\mathrm{A} = \mathcal{A} \cup \{a_\mathrm{LQR}\},
\end{equation}

where the abstract action $a_\mathrm{LQR}$, when selected, applies the LQR control. Due to the use of an abstract action, this method is only applicable to discrete action set methods such as DQN. The resulting \emph{mixed cartesian-abstract} action set can be represented by adding an integer-valued abstract action dimension to the action space, where the value $0$ indicates the use of the cartesian action space and any other value (in this case, only 1) indicates the index of the abstract action.

\subsection{Integrated LQR Action}

While the abstract LQR action does not share the cartesian action space with the other actions, the applied LQR control does. We may therefore instead add the LQR control as a sample point in the original action space by making it state-dependent

\begin{equation}
\mathcal{A}^\mathrm{IA}(s) = \mathcal{A} \cup \{-F\overline{s} + a_\mathrm{ff}\}
\end{equation}

and taking the maxium over $\mathcal{A}^\mathrm{IA}(s)$ instead of over $\mathcal{A}$ in Eqs~\ref{eq:loss} and~\ref{eq:egreedy}. The advantage of this method over an abstract LQR action is that its return may be generalized over by the Q-value representation. Taking similar (discrete) actions in the regular action space therefore also improves the estimate of the LQR action, decreasing the learning time.

In addition, we can embed the LQR action in the DDPG algorithm by taking the $\argmax$ in Eqs~\ref{eq:loss} and~\ref{eq:egreedy} over both the LQR action $-F\overline{s}+ a_\mathrm{ff}$ and the DDPG action $\mu(s|\theta^{\mu'})$. While DDPG can already stabilize a dynamic system without chatter, adding a known-good (at least close to the goal state) solution might also decrease the learning time, or increase performance. To ensure sufficient exploration, we use $\epsilon$-greedy exploration for the choice between DDPG and LQR action, and apply the exploration noise in Eq.~\ref{eq:gauss} to both.

\section{Simulations}
\label{sec:experiments}

We test the three described methods on three simulated testbeds: the pendulum swing-up, cart-pole swing-up and 2d flyer, using the Generic Reinforcement Learning Library, GRL\footnote{Code available at \url{https://github.com/wcaarls/grl}.}. The network architecture and other parameters are described in Appendix~\ref{sec:parameters}.

All experiments were performed 20 times to calculate the 95\% confidence interval of the results, which are presented as the rise time and end performance. The rise time is defined as the first time the agent passes a system-defined cumulative episode reward consistently (3 times in a row), and the end performance is the mean of the cumulative episode reward over the last 10\% of the episodes of each run. 

\subsection{Pendulum Swing-up}

The pendulum swing-up is a classical control problem where a pendulum attached to a motor has to swing up from the stable equilibrium to the unstable equilibrium, but without enough torque to do so in one swing~(\cite{bbse:rlfa}). The system has two state dimensions $s=[\theta, \dot{\theta}]$ being the angle and angular velocity of the pendulum and one action dimension $a=u$ being the voltage applied to the motor, up to 3$V$. To avoid the $0/2\pi$ nonlinearity, the angle $\theta$ is supplied to the networks in a sine-cosine representation. The cost matrices are $C=\mathrm{diag}([5, 0.01])$ and $D=1$, with goal state $s_\mathrm{d}=[0, 0]$. The episode ends after $3$s.

\subsection{Cart-pole Swing-up}

Another classical control set-up is the cart-pole swing-up~(\cite{bsa:control}). In this case, the pendulum is mounted to a cart, which can be pushed along a track to perform the swing-up. As such, the system has four state dimensions $s=[x,\theta,\dot{x},\dot{\theta}]$, now including the position and velocity of the cart. The action is still one-dimensional $a=F$, being the force applied to the cart, up to $15$N. Again, the angle is presented in sine-cosine representation, and the cost matrices are $C=\mathrm{diag}(2, 1, 0.1, 0.1)$ and $D=\frac{1}{15}$, with goal state $s_\mathrm{d}=[0, 0, 0, 0]$. The episode ends after $10$s.

\subsection{2d Flyer}

We introduce the 2d flyer as a very unstable regulation task, with additional nonlinearity in the form of an obstacle. The flyer is modeled as a rod with mass $m=0.1$kg and length $l=0.1$m, where two forces $[F_\mathrm{L}, F_\mathrm{R}]$ perpendicular to the rod may be applied at the tips. The equations of motion are

\begin{equation}
\left(\begin{matrix}
\ddot{x}\\
\ddot{y}\\
\ddot{\theta}\\
\end{matrix}\right)
= \left(\begin{matrix}
-\left(F_\mathrm{L}+F_\mathrm{R}\right)\sin\theta/m\\
\left(F_\mathrm{L}+F_\mathrm{R}\right)\cos\theta/m-g\\
\left(F_\mathrm{R}-F_\mathrm{L}\right)l/I\\
\end{matrix}\right),
\end{equation}

where $g=9.81$ and the inertia $I=\frac{ml^2}{3}$. Therefore, the system has six state dimensions $s=[x, y, \theta, \dot{x}, \dot{y}, \dot{\theta}]$ and two action dimensions $a=[F_\mathrm{L}, F_\mathrm{R}]-[0.5, 0.5]$, up to 0.1N. The angle is represented by its sine and cosine once again, and the cost matrices are $C=\mathrm{diag}([1,1,1,0,0,0])$ and $D=\mathrm{diag}([1, 1])$, with goal state $s_\mathrm{d}=[0, 0, 0 ,0, 0 ,0]$. The episode ends after $20$s or if the flyer leaves the target area $[-1, -1] < [x, y] < [1, 1]$. 

An obstacle occupies the region $[-0.4, -0.3] < [x, y] < [0.1, -0.2]$, which prohibits the flyer from reaching the target through simple regulation from the start location $[-0.4, -0.4]$.

\section{Results}
\label{sec:results}

\begin{figure*}
\centering
\subfloat[Pendulum swing-up DQN]{
\includegraphics[width=0.45\linewidth]{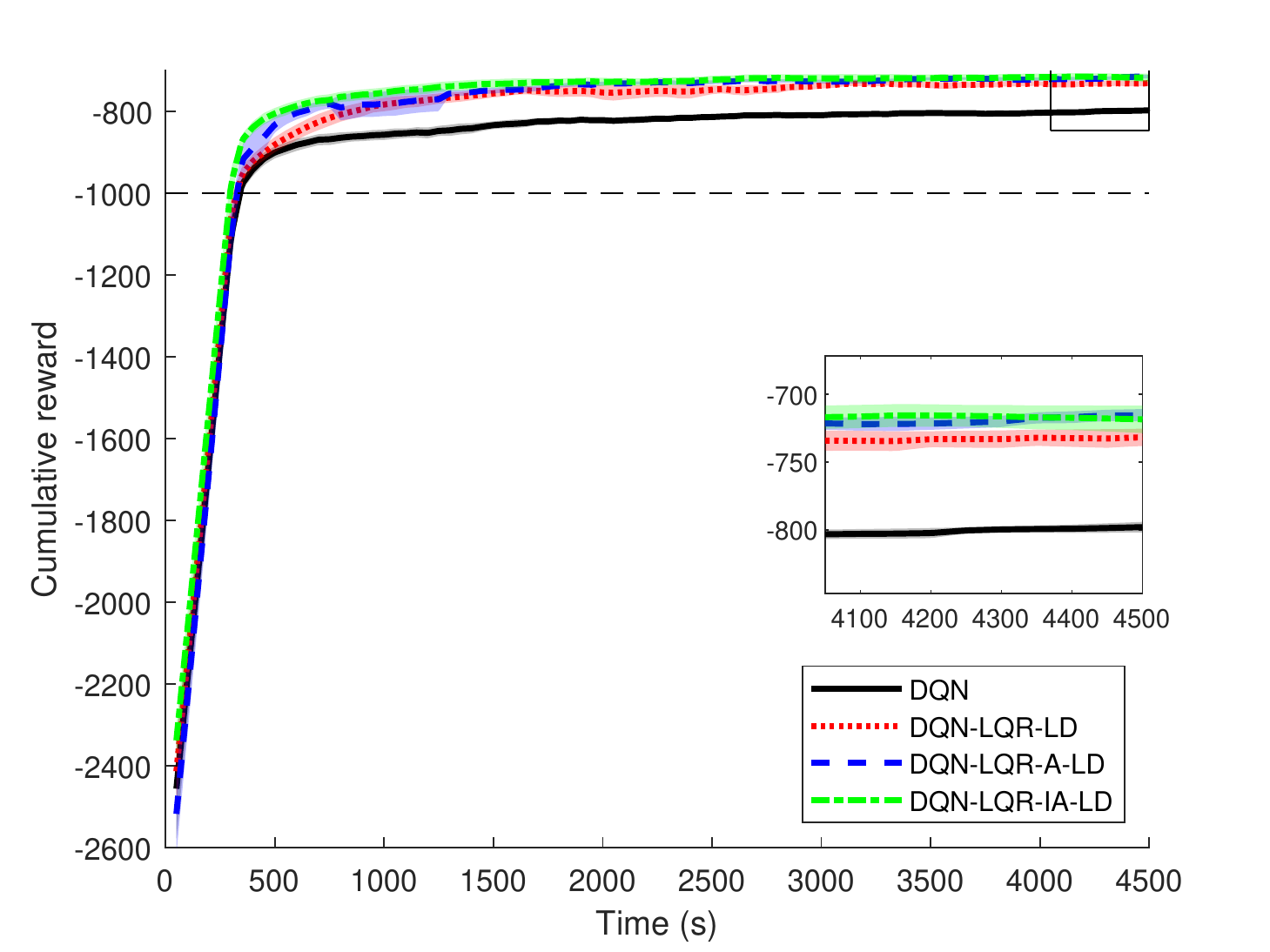}
}
\hfil
\subfloat[Pendulum swing-up DDPG]{
\includegraphics[width=0.45\linewidth]{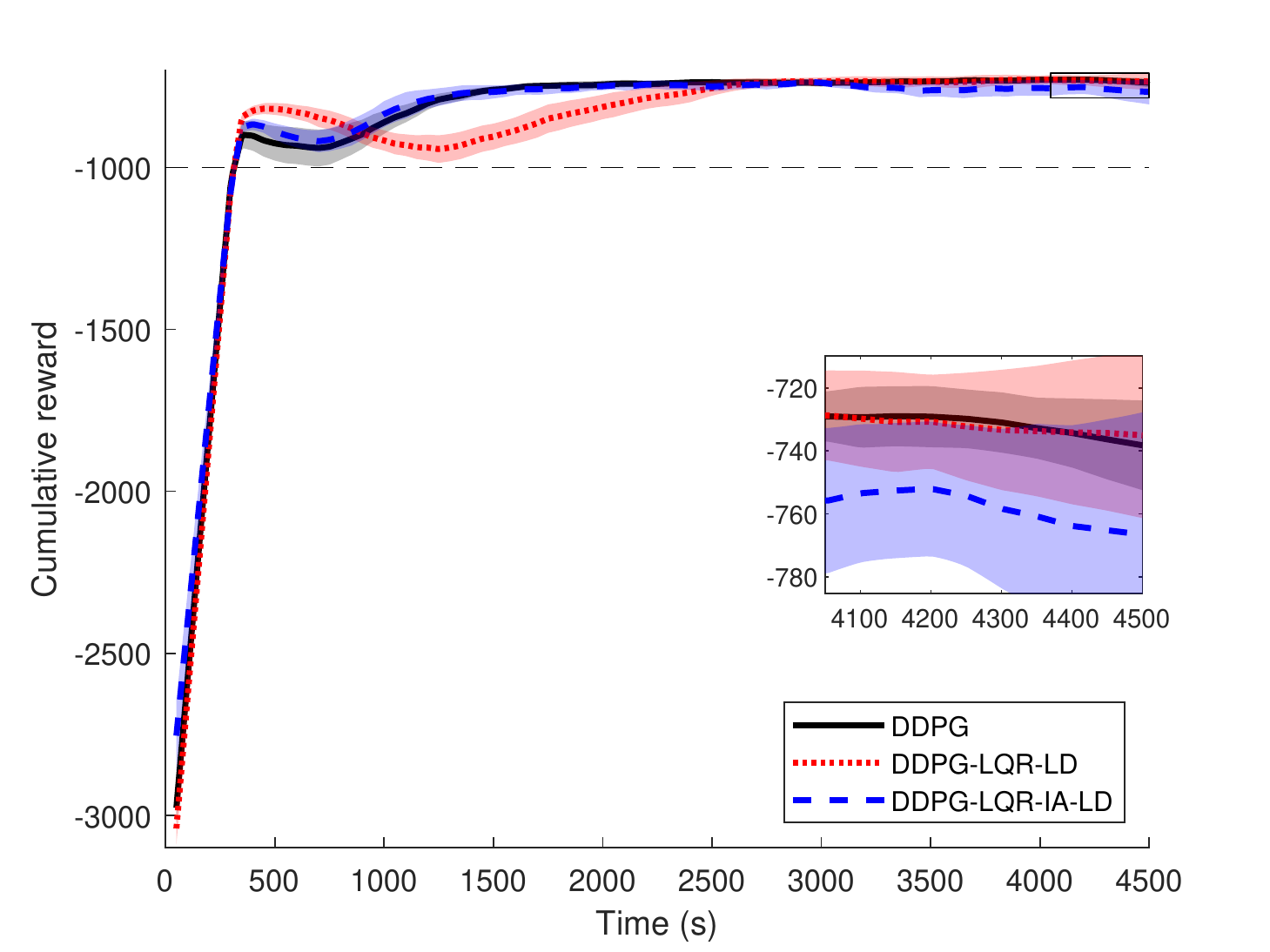}
}\\
\centering
\subfloat[Cart-pole swing-up DQN]{
\includegraphics[width=0.45\linewidth]{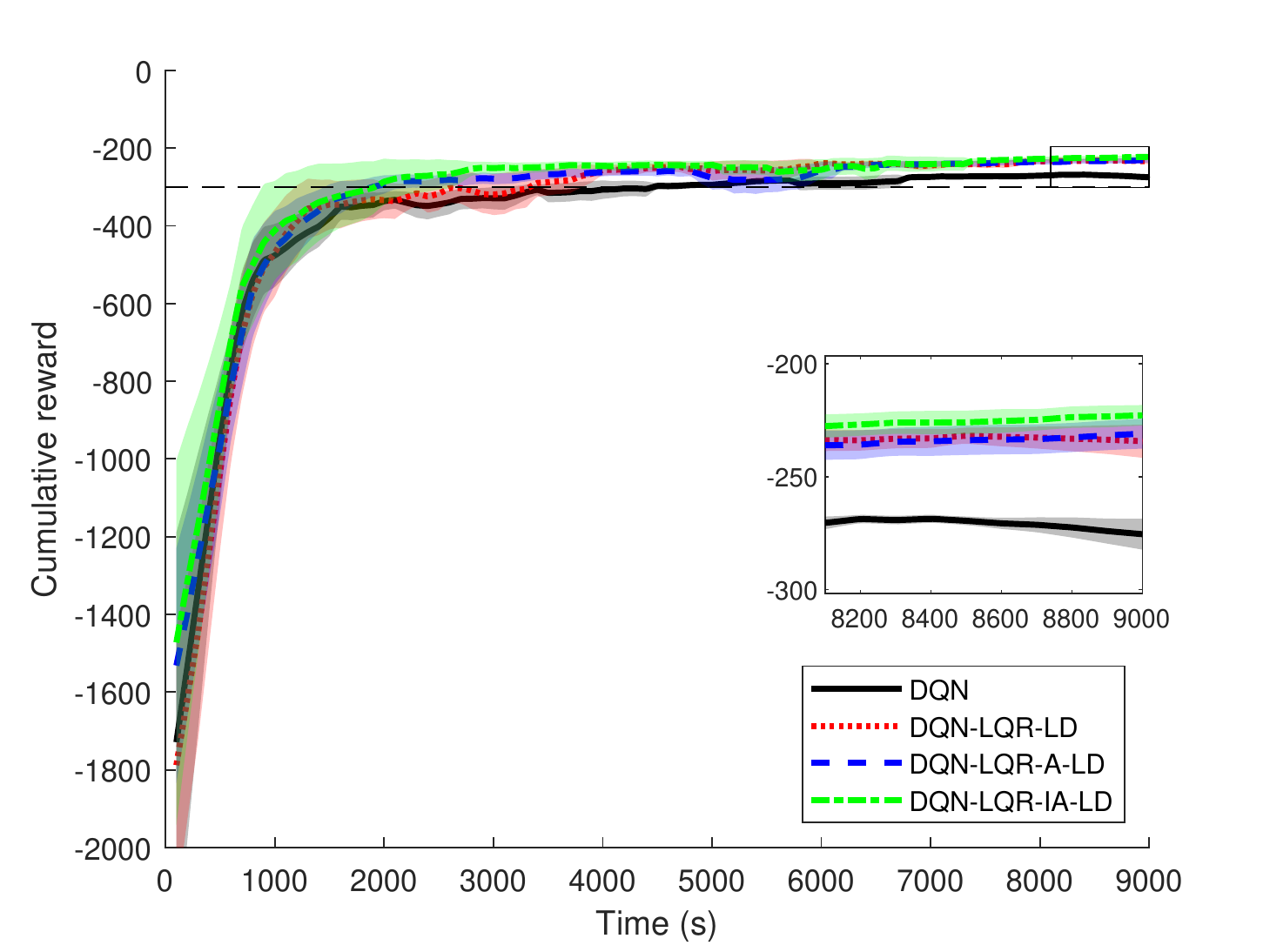}
}
\hfil
\subfloat[Cart-pole swing-up DDPG]{
\includegraphics[width=0.45\linewidth]{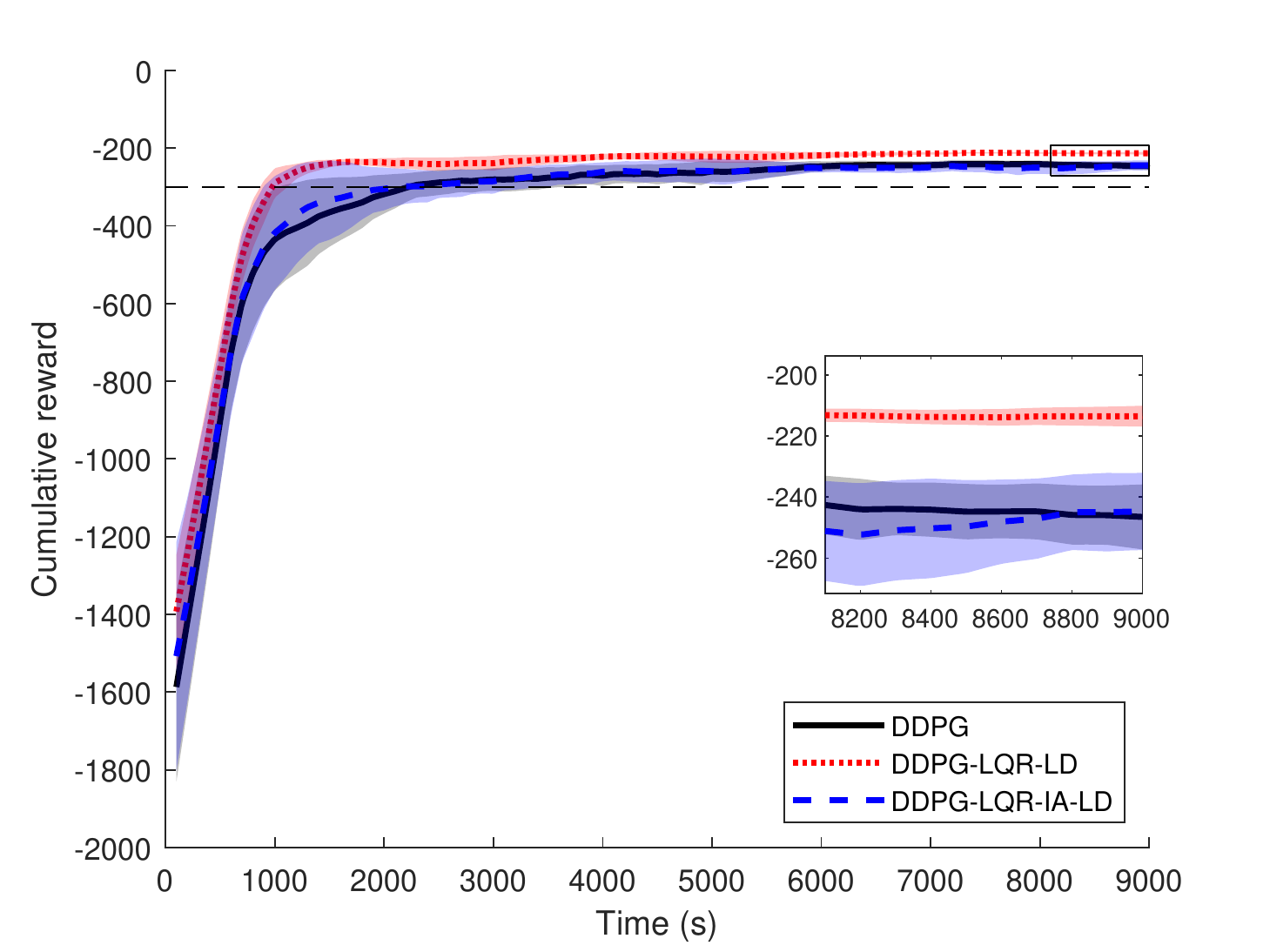}
}\\
\centering
\subfloat[2d flyer DQN]{
\includegraphics[width=0.45\linewidth]{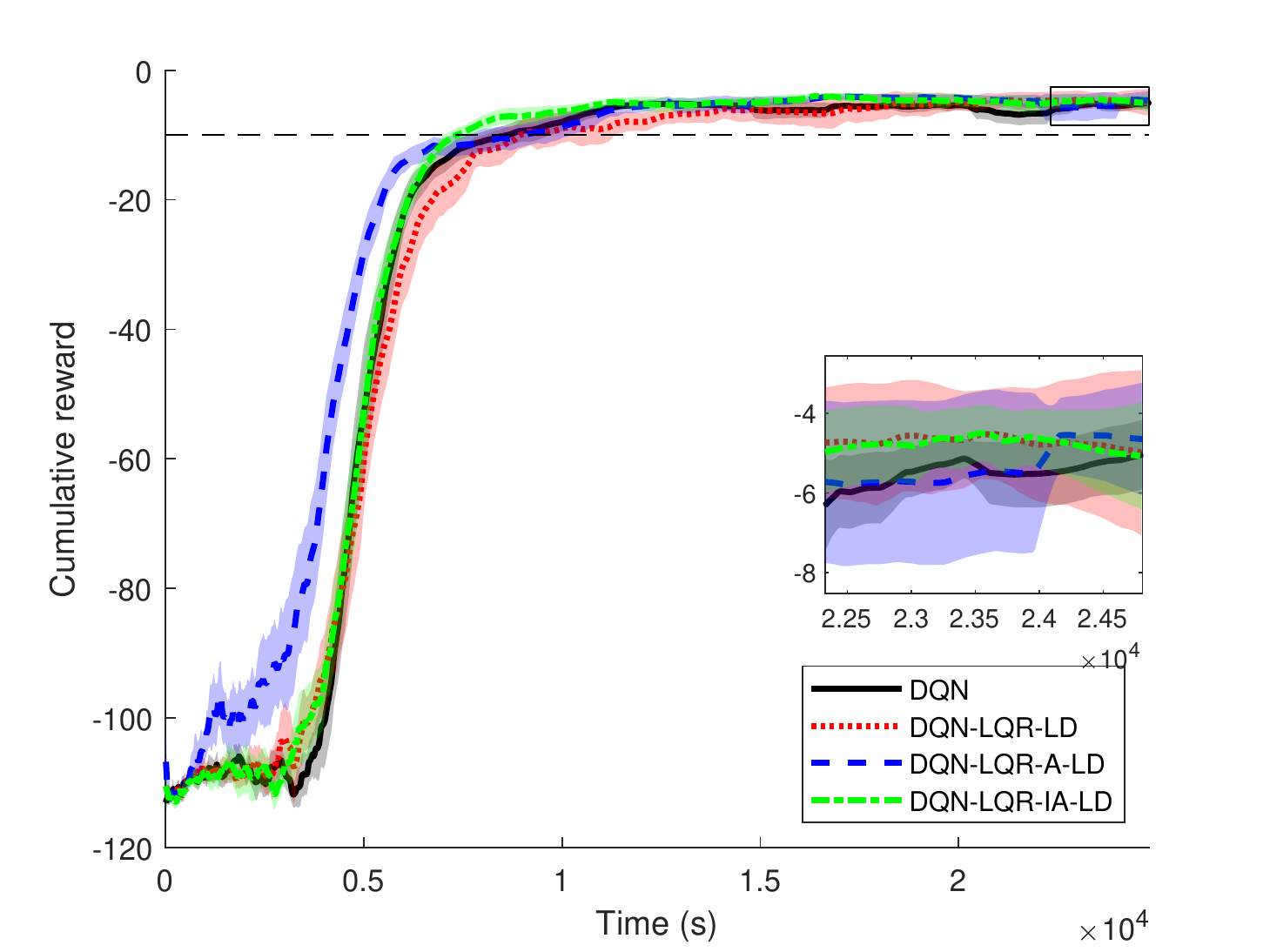}
}
\hfil
\subfloat[2d flyer DDPG]{
\includegraphics[width=0.45\linewidth]{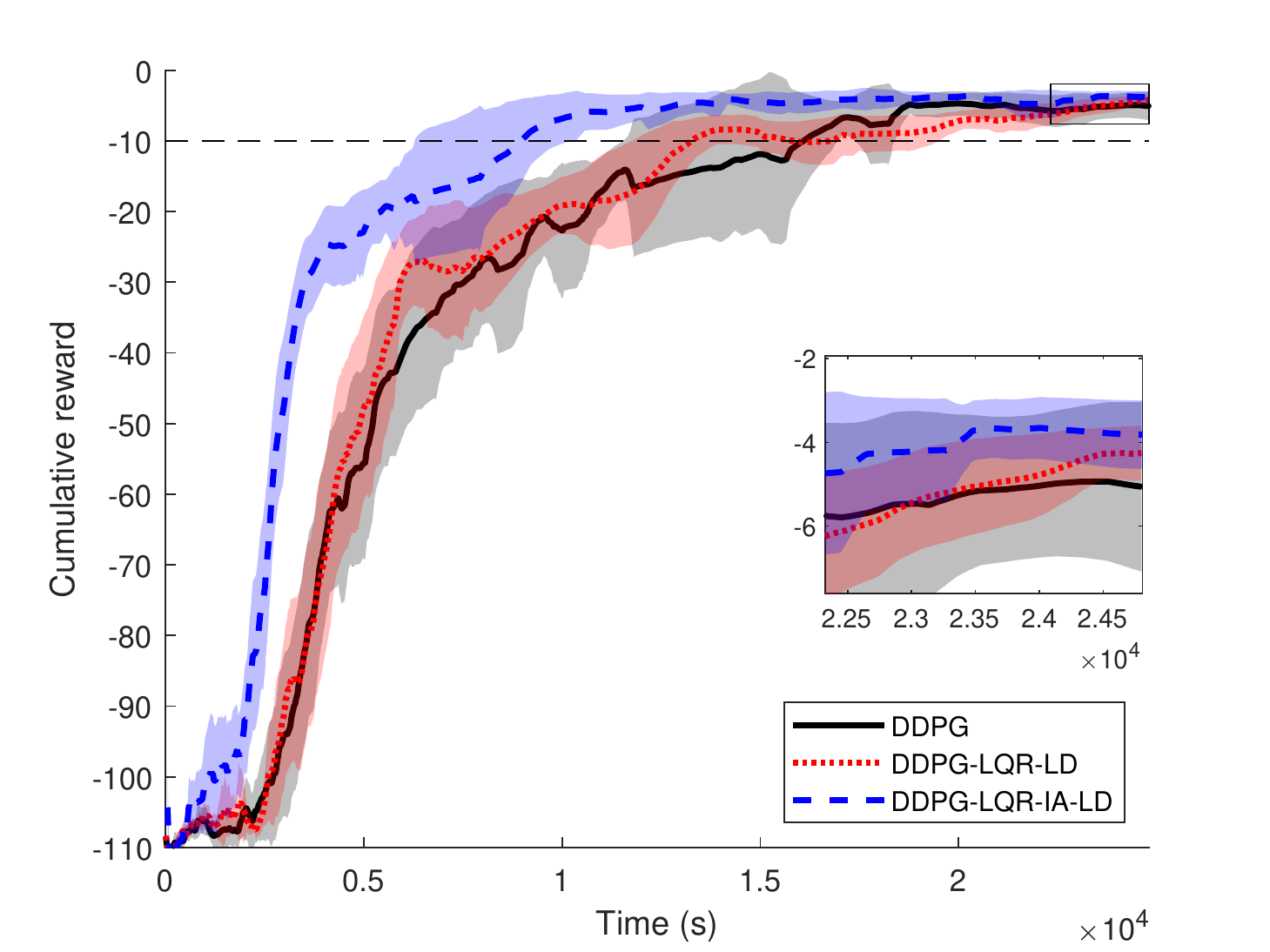}
}
\caption{Performance evaluation of LQR integration using learned dynamics for DQN (left column) and DDPG (right column). Shown are the mean and 95\% confidence interval over 20 independent runs, plotted using a moving average filter of 10 episodes. The horizontal line is the point at which the rise time is measured.}
\label{fig:results:curves}
\end{figure*}

The main results are presented in Table~\ref{tab:results}, with the respective learning curves given in Figure~\ref{fig:results:curves}.

\begin{table*}
\caption{Mean and 95\% confidence intervals over 20 runs for the rise time and end performance on the three testbeds. Values in bold are the best performance within that category (DQN or DDPG), or statistically equivalent to it.}
\label{tab:results}
\begin{tabular}{lrp{0em}rrp{0em}rrp{0em}rrp{0em}rrp{0em}rrp{0em}r}
\hline
&\multicolumn{6}{c}{pendulum swing-up}&\multicolumn{6}{c}{cart-pole swing-up}&\multicolumn{6}{c}{2d flyer}\\
&\multicolumn{3}{c}{rise (s)}&\multicolumn{3}{c}{end perf}&\multicolumn{3}{c}{rise (s)}&\multicolumn{3}{c}{end perf}&\multicolumn{3}{c}{rise (s)}&\multicolumn{3}{c}{end perf}\\
\hline
DQN&350&$\pm$&46&-802&$\pm$&3.5&2405&$\pm$&536&-270&$\pm$&1.7&7749&$\pm$&561&-5.4&$\pm$&0.8\\
DQN-LQR&313&$\pm$&40&-736&$\pm$&9.1&\textbf{985}&$\pm$&191&-232&$\pm$&5.0&7563&$\pm$&804&-6.9&$\pm$&2.0\\
DQN-LQR-A&285&$\pm$&31&\textbf{-714}&$\pm$&2.9&1625&$\pm$&401&-236&$\pm$&6.8&\textbf{6348}&$\pm$&747&\textbf{-4.3}&$\pm$&0.7\\
DQN-LQR-IA&\textbf{245}&$\pm$&36&\textbf{-713}&$\pm$&1.9&1370&$\pm$&352&\textbf{-224}&$\pm$&3.7&\textbf{7012}&$\pm$&700&\textbf{-4.3}&$\pm$&0.5\\
DQN-LQR-LD&308&$\pm$&42&-733&$\pm$&5.7&1210&$\pm$&337&-233&$\pm$&4.3&7103&$\pm$&442&\textbf{-4.5}&$\pm$&1.2\\
DQN-LQR-A-LD&320&$\pm$&52&-721&$\pm$&4.3&1900&$\pm$&410&-234&$\pm$&6.3&\textbf{6784}&$\pm$&609&-5.4&$\pm$&1.8\\
DQN-LQR-IA-LD&\textbf{233}&$\pm$&31&-716&$\pm$&8.3&1280&$\pm$&418&\textbf{-226}&$\pm$&5.1&7202&$\pm$&439&\textbf{-4.7}&$\pm$&0.7\\
\hline
DDPG&213&$\pm$&67&\textbf{-731}&$\pm$&10&1590&$\pm$&600&-244&$\pm$&8.8&10993&$\pm$&1493&-5.1&$\pm$&1.7\\
DDPG-LQR&\textbf{185}&$\pm$&20&\textbf{-728}&$\pm$&12&1025&$\pm$&351&-216&$\pm$&4.9&8840&$\pm$&1585&-4.6&$\pm$&1.7\\
DDPG-LQR-IA&\textbf{198}&$\pm$&15&\textbf{-737}&$\pm$&10&1755&$\pm$&559&-264&$\pm$&29&\textbf{5470}&$\pm$&841&\textbf{-3.4}&$\pm$&0.4\\
DDPG-LQR-LD&\textbf{183}&$\pm$&16&\textbf{-731}&$\pm$&16&\textbf{805}&$\pm$&124&\textbf{-214}&$\pm$&2.4&8314&$\pm$&1327&-4.9&$\pm$&1.0\\
DDPG-LQR-IA-LD&\textbf{198}&$\pm$&13&-754&$\pm$&22&1310&$\pm$&531&-249&$\pm$&15&\textbf{6140}&$\pm$&912&\textbf{-3.7}&$\pm$&0.7\\
\end{tabular}

\end{table*}

\begin{figure*}
\centering
\subfloat[2d flyer state evolution, DQN]{
\includegraphics[width=0.45\linewidth]{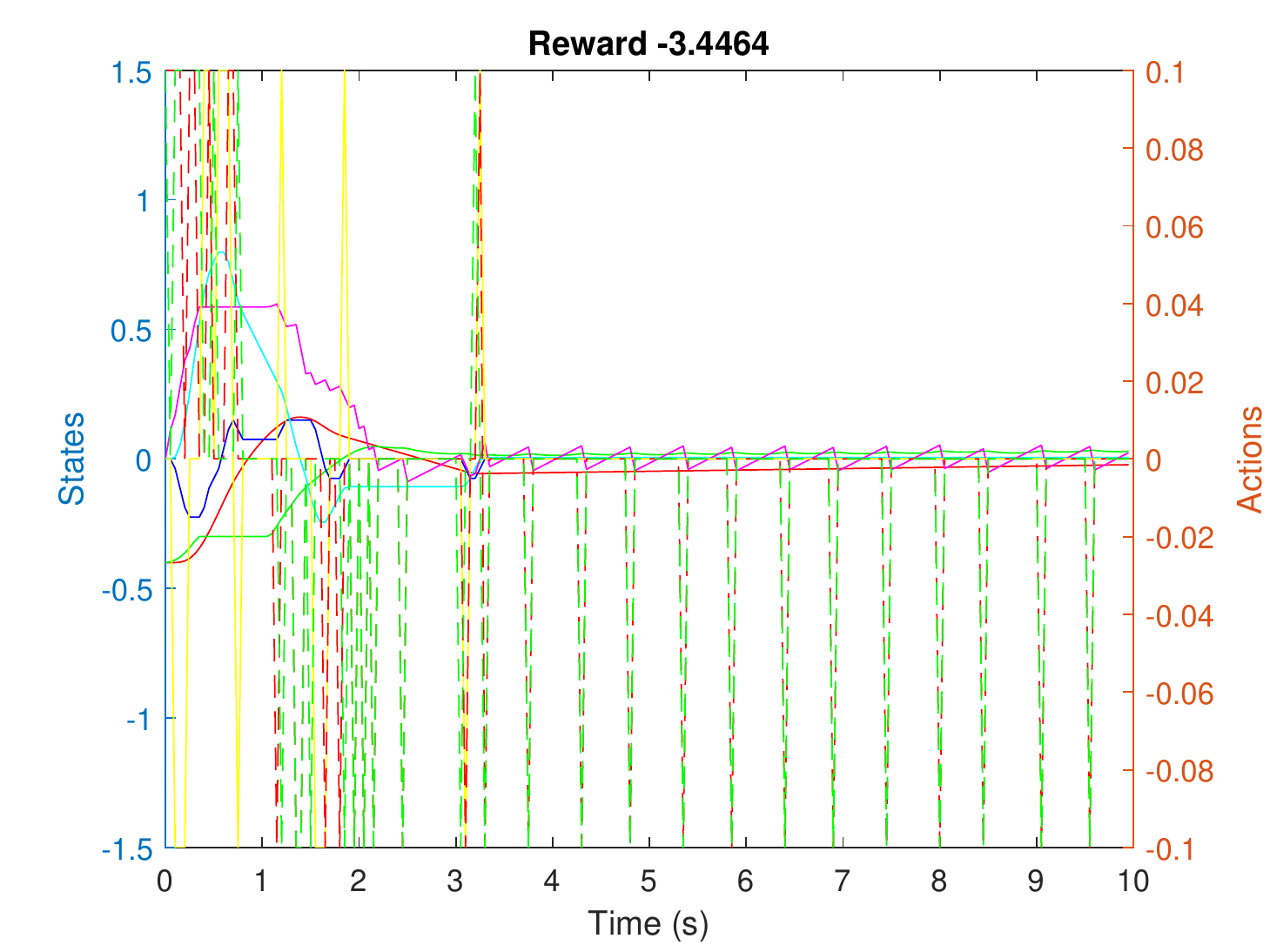}
}
\hfil
\subfloat[2d flyer state evolution, DQN with integrated LQR action]{
\includegraphics[width=0.45\linewidth]{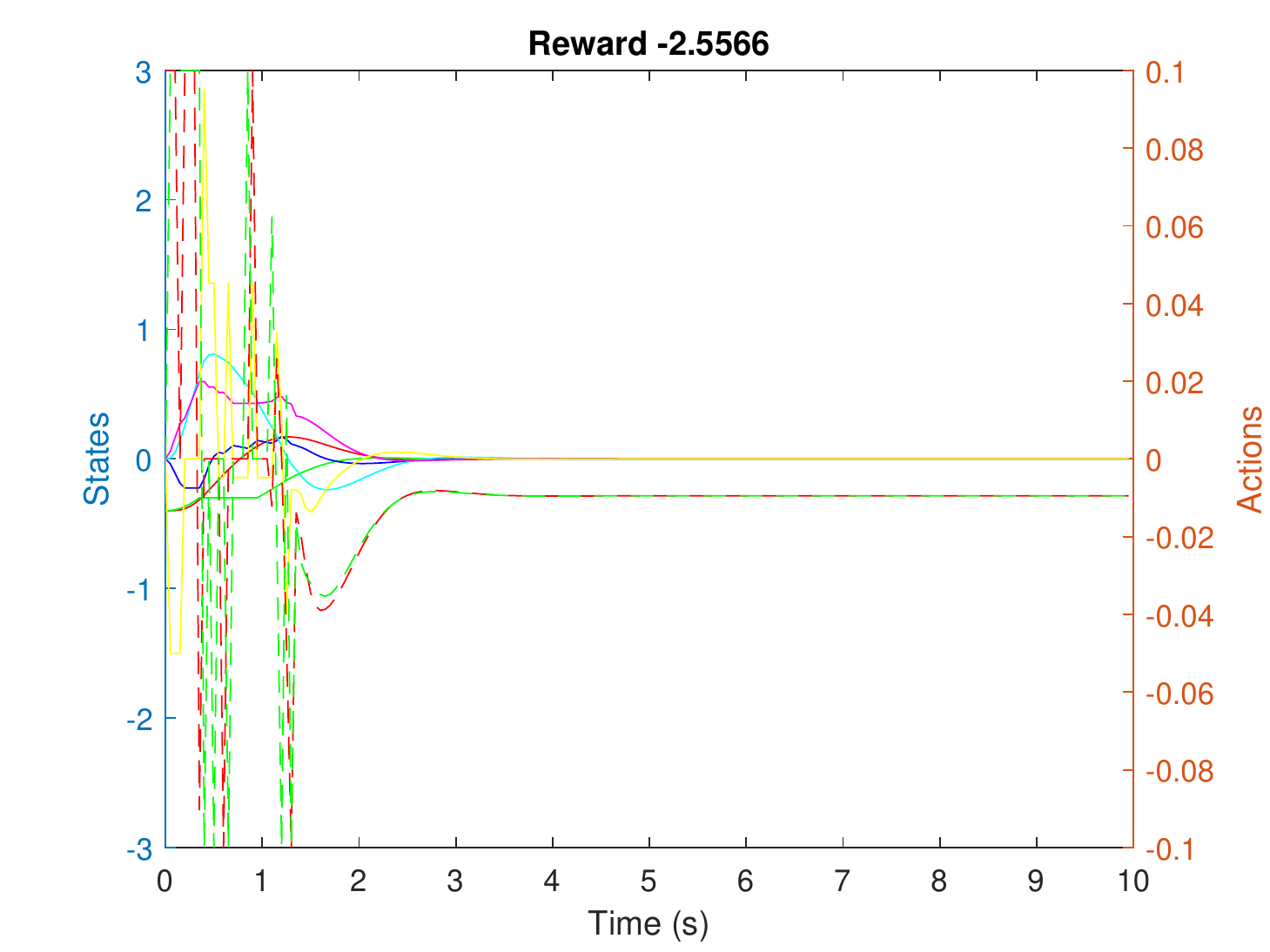}
}
\caption{State evolution of DQN and DQN-LQR-IA. Solid lines are state values, dashed lines are actions. DQN suffers from chattering due to its discrete action set, while DQN-LQR-IA maintains a chatter-free equilibrium.}
\label{fig:results:state}
\label{fig:results:evolution_dqn}
\end{figure*}

\begin{figure*}
\centering
\subfloat[2d flyer state evolution, DDPG]{
\includegraphics[width=0.45\linewidth]{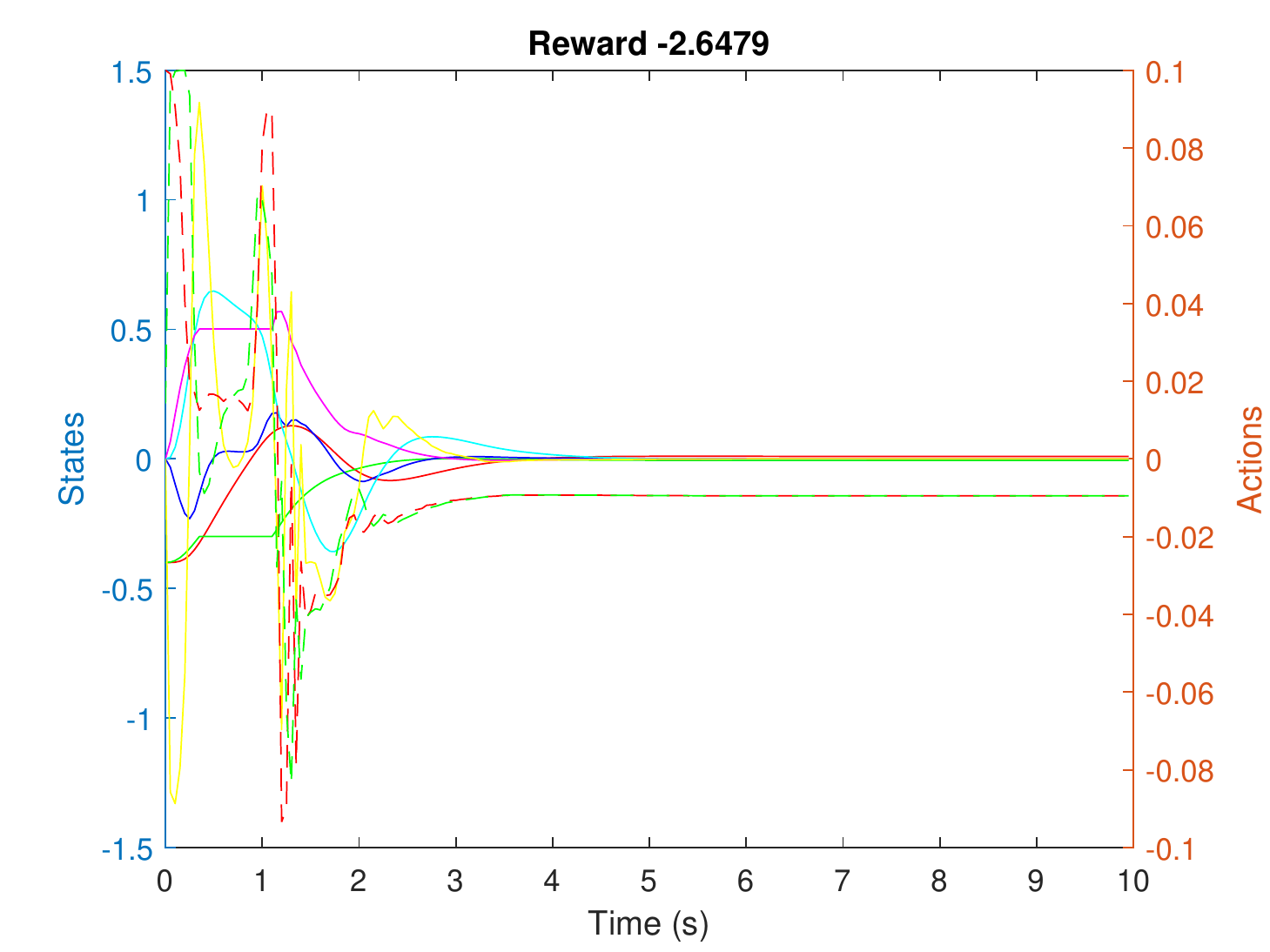}
}
\hfil
\subfloat[2d flyer state evolution, DDPG with integrated LQR action]{
\includegraphics[width=0.45\linewidth]{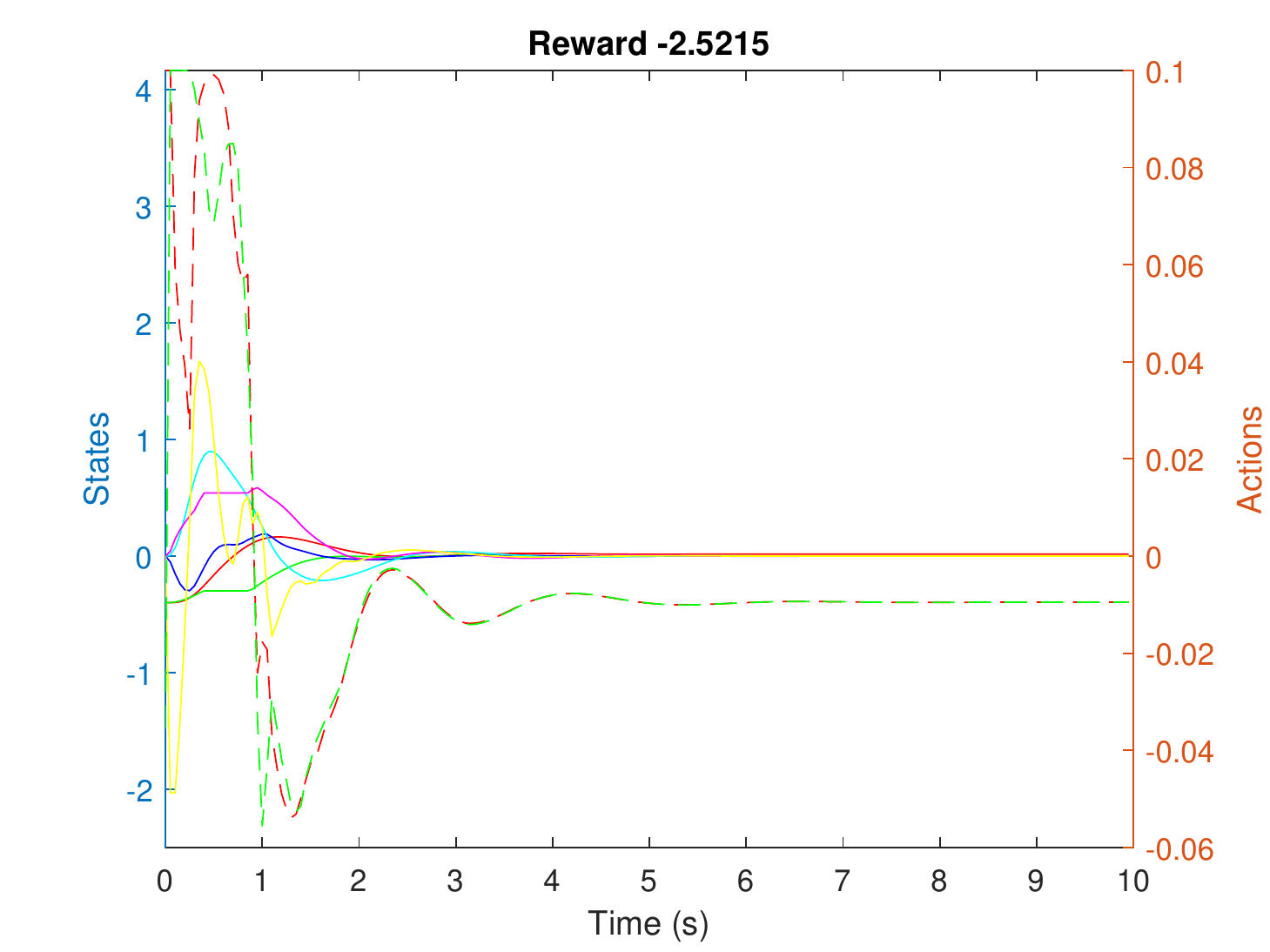}
}
\caption{State evolution of DDPG and DDPG-LQR-IA. Solid lines are state values, dashed lines are actions. DDPG-LQR-IA has a lower steady-state error.}
\label{fig:results:evolution_ddpg}
\end{figure*}

\subsection{DQN}

In the DQN case, the LQR integrated action algorithm (DQN-LQR-IA) consistently presented the best end performance, although it did not learn significantly faster than using an abstract action (DQN-LQR-A). All LQR-based algorithms outperform baseline DQN due to reduced chattering while maintaining the goal state, see Figure~\ref{fig:results:evolution_dqn}.

Using learned dynamics (*-LD) did not have a large effect on either rise time or end performance, indicating that the dynamics around the goal state are learned sufficiently quickly such as not to impede the learning process. In theory, learned dynamics makes the system nonstationary, as the actions of the LQR controller change during the run. The impact of this nonstationarity is different for each system. For DQN-LQR-LD, it changes the rewards received when entering the capture region (see Eq.~\ref{eq:smdp}), while for DNQ-LQR-A-LD it changes the controls applied by the abstract action, leading to a different next state. Finally, for DQN-LQR-IA-LD, it changes the action set used to calculate the target values. Note, however, that while these errors are stored permanently in the replay memory in the case of DQN-LQR-LD and DQN-LQR-A-LD, for DQN-LQR-IA-LD the loss in Eq.~\ref{eq:loss} is recalculated every time a new minibatch is sampled, and thus always reflects the current dynamics.

\subsection{DDPG}

Because DDPG already uses continuous actions, we did not expect a large performance gain from using LQR actions. Indeed, for the pendulum plain DDPG has the best end performance and is only slightly slower than the other variants. However, LQR capture (DDPG-LQR(-LD)) shows significantly improved rise time and end performance for the cart-pole swing-up problem, while LQR integrated action (DDPG-LQR-IA(-LD)) has the best rise time and end performance for the 2d flyer.

The improvements in rise time and end performance for LQR capture show that using a well-chosen capture region in which the LQR controller is optimal helps learning, even when using a continuous action algorithm such as DDPG. And the good results for LQR integrated action for the 2d flyer indicate that choosing between the LQR and DDPG actions might help guide the solution towards a better policy, although similar to (\cite{Gu2016mba}) the result is inconsistent across domains.

Inspecting the state evolution of individual episodes during a run (data not shown) shows that the best episode reward is comparable for all methods. Rather, it is the average end performance that is improved.

\section{Conclusion}
\label{sec:conclusion}

We presented a brief comparison between three different ways of using the action calculated by an optimal controller in reinforcement learning: LQR capture, LQR action and LQR integrated action. When combined with DQN, all methods decreased rise time and increased end performance compared to the baseline, with the LQR integrated action algorithm having the best performance overall. For DDPG the results are less consistent, but both LQR capture and LQR integrated action showed improved performance in some test cases.

Future work includes using different (non-linear) optimal control techniques instead of LQR, which allows tackling non-regulation tasks. In this setting, it would also be interesting to compare our approach to methods that somehow combine the control output with the RL policy, such as by summing (\cite{kkvbc:mismatch}). The fact that LQR integrated action does not poison the replay memory may be particularly advantageous when integrating learning or adaptive controllers. 

\bibliography{root}

\appendix

\section{Parameters}
\label{sec:parameters}

Table~\ref{tab:netparam} contains the environment-independent configuration of the DQN and DDPG algorithms, while Table~\ref{tab:envparam} contains the environment-specific parameters.

\begin{table}[h!]
\centering
\caption{Algorithmic parameters}
\label{tab:netparam}
\begin{tabular}{lcc}
\hline
Parameter&DQN&DDPG\\
\hline
Hidden layers&[400, 300]&[400, 300]\\
Hidden layer activation&ReLU&ReLU\\
Output activation&linear&tanh\\
\hline
Exploration rate ($\epsilon$)&0.05&\\
\cite{ou:noise} friction&&0.15\\
\end{tabular}
\end{table}

\begin{table}[h!]
\centering
\caption{Environment-specific parameters}
\label{tab:envparam}
\begin{tabular}{lccc}
\hline
Parameter&pendulum&cart-pole&2d flyer\\
\hline
State dimensions&2&4&6\\
Action dimensions&1&1&2\\
Action discretization&3&3&[3, 3]\\
Control time step ($\tau$)&0.03&0.05&0.05\\
Timeout&3s&10s&20s\\
\hline
Discount rate ($\gamma$)&0.99&0.97&0.99\\
Exploration noise ($\sigma$)&1&5&0.01\\
Reward scale&0.1&0.1&1\\
Replay memory&$\infty$&$\infty$&$\infty$\\
\hline
LLR neighbors ($K$)&64&64&64\\
LLR memory&10000&10000&10000\\
\end{tabular}
\end{table}

\end{document}